\documentclass{article}
\usepackage{ijcai23}

\usepackage{times}
\usepackage{soul}
\usepackage{url}
\usepackage[hidelinks]{hyperref}
\usepackage[utf8]{inputenc}
\usepackage[small]{caption}
\usepackage{graphicx}
\usepackage{amsmath}
\usepackage{amsthm}
\usepackage{booktabs}
\usepackage{algorithm}
\usepackage{algorithmic}
\usepackage[switch]{lineno}
\usepackage{xcolor}         % colors
\title{Learning Monocular Depth in Dynamic Environment via \\ Context-aware Temporal Attention}
\author{
Zizhang Wu$^{1}$\footnote{These authors contributed equally.}
\and
Zhuozheng Li$^{1}$$^{*}$
\and
Zhi-Gang Fan$^{1}$$^{*}$
\and
Yunzhe Wu $^1$
\and
Yuanzhu Gan$^1$
\\
Jian Pu$^2$
\and
Xianzhi Li$^3$
\affiliations
$^1$ZongmuTech\and
$^2$Fudan University\and
$^3$Huazhong University of Science and Technology\\
\emails
wuzizhang87@gmail.com,
\{zhuozheng.li, zhigang.fan, nelson.wu, yuanzhu.gan\}@zongmutech.com\\
jianpu@fudan.edu.cn\and
xzli@hust.edu.cn
}

\begin{document}

\maketitle

\begin{abstract}
    The monocular depth estimation task has recently revealed encouraging prospects, especially for the autonomous driving task. To tackle the ill-posed problem of 3D geometric reasoning from 2D monocular images, multi-frame monocular methods are developed to leverage the perspective correlation information from sequential temporal frames. However, moving objects such as cars and trains usually violate the static scene assumption, leading to feature inconsistency deviation and misaligned cost values, which would mislead the optimization algorithm. In this work, we present \textbf{CTA-Depth}, a \textbf{C}ontext-aware \textbf{T}emporal \textbf{A}ttention guided network for multi-frame monocular \textbf{Depth} estimation. Specifically, we first apply a multi-level attention enhancement module to integrate multi-level image features to obtain an initial depth and pose estimation. Then the proposed CTA-Refiner is adopted to alternatively optimize the depth and pose. During the refinement process, context-aware temporal attention (CTA) is developed to capture the global temporal-context correlations to maintain the feature consistency and estimation integrity of moving objects. In particular, we propose a long-range geometry embedding (LGE) module to produce a long-range temporal geometry prior. Our approach achieves significant improvements over state-of-the-art approaches on three benchmark datasets. 
    % (e.g., 13.5\% for Abs Rel on the KITTI dataset) We will release our code for implementation after paper acceptance.
\end{abstract}

\section{Introduction}

Monocular depth estimation aims at predicting accurate pixel-wise depth from monocular RGB images.
Due to its low cost and easy implementation, monocular depth estimation has achieved promising prospects in practical applications~\cite{li2022densely,9922174,mumuni2022bayesian}.
Particularly, monocular depth estimation~\cite{li2015depth,liu2015deep,ricci2018monocular,bhat2021adabins,yuan2022newcrfs} under the single-frame setting has achieved convincing results by conducting robust convolutional neural networks with prior geometric constraints.
Nevertheless, it is still challenging to precisely recover the 3D environment from a single monocular 2D image. 
On the other hand, noting that sequential image frames are achievable from the monocular camera, existing studies~\cite{wang2019recurrent,zhang2019exploiting,patil2020don} start paying greater attention to the depth estimation under the multi-frame setting. 

Inspired by the stereo matching task, multi-frame monocular depth estimation works typically employed the cost volume or cost map~\cite{watson2021temporal,gu2023dro,bae2022multi} to accomplish geometric reasoning and have gradually achieved state-of-the-art performance.
% succeeded in achieving the state-of-the-art.see regions marked by white arrows.
\begin{figure}[t]      
    \centering
    \includegraphics[width =0.48\textwidth]{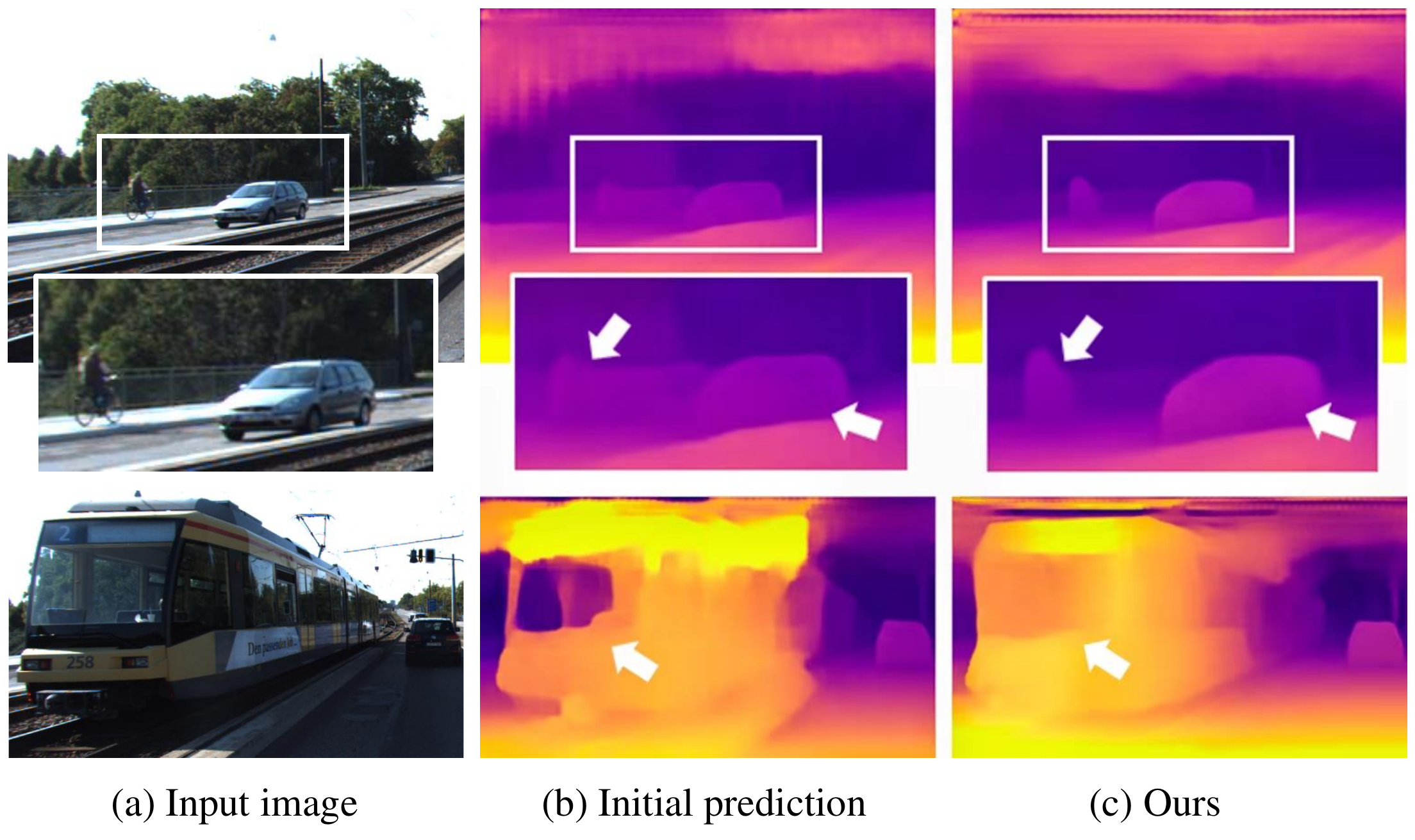}
    \vspace{-2mm}
    \caption{Given input images, our CTA-Depth with CTA-Refiner predicts more accurate depth maps compared to the initial prediction, especially for dynamic objects.
    }
    \vspace{-4mm}
    \label{fig:teaser}
\end{figure}

However, the widely applied static scene assumption \cite{klingner2020self,li2019learning} for the construction of the cost volume does not always hold in real-world scenarios.
Specifically, moving objects such as cars, trains and pedestrians result in feature inconsistency deviation, misaligned cost values and degraded re-projection loss, which would mislead the optimization algorithm. 
To address this issue, recent works~\cite{lee2021learning,feng2022disentangling,wimbauer2021monorec} attempted to solve dynamic problems by introducing an auxiliary object motion prediction module and segmentation masks to predict or disentangle dynamic objects explicitly.
It inevitably increases the complexity and redundancy of the model and ignores the sustained temporal relation modeling of moving objects across long-range multiple frames, which thus limits the potential of sequential images for time-crucial industry implementation. 

%They lack the effective prior for dynamic objects to capture the moving objects of long-range frames.
%it is under-estimated with the expert knowledge for learning temporal geometric constraints at the feature level and potential of the recurrent architecture, which limits the further exploration specifically for the dynamic objects in the depth estimation task.
%learn temporal geometric constraints between moving objects

Thus, to efficiently boost the multi-frame context-temporal feature integration for dynamic targets without explicit auxiliary modules, we propose our \textbf{CTA-Depth}, a \textbf{C}ontext-aware \textbf{T}emporal \textbf{A}ttention network for joint multi-frame monocular \textbf{Depth} and pose estimation.
%which avoids the explicit object motion prediction or disentanglement module.
Specifically, we first utilize the multi-level attention enhancement (MAE) module for reliable initial estimation, which applies cross-scale attention layers to achieve ample interaction of different-scale features to equip the network with both local and global attentions.
%-------------------
Furthermore, we develop the refiner CTA-Refiner to iteratively optimize our predictions with the inputs of context and temporal features. In specific, we develop our depth-pose context-aware temporal attention (CTA) with the cross-attention mechanism that assigns the temporal features as values, and the context features as queries and keys.
As a result, it implicitly interacts with the context and temporal features to maintain the estimation integrity of moving objects across multiple sample frames.
Additionally, to expand the temporal field of interest and aggregate useful clues for geometry reasoning, especially for dynamic targets within distant frames, we present our long-range geometry embedding (LGE) and provide it to the CTA process for seizing the long-range temporal geometry prior.

Qualitatively, as shown in Fig.~\ref{fig:teaser}(b), it is difficult for the single-frame method to recover the complete geometry of moving objects, while the predictions from our implicit CTA-Depth in Fig.~\ref{fig:teaser}(c) demonstrate its robustness at high-level feature recognition. Quantitatively, we conduct extensive experiments on three challenging benchmarks to validate the effectiveness of our pipeline against state-of-the-art models. 
% Please see the experiments section for more detailed comparisons.
We summarize our contributions as follows:
\begin{itemize}
    \item  We propose our CTA-Depth, an implicit, long-range Context-aware Temporal Attention guided network for supervised multi-frame monocular depth estimation, focusing on the dynamic object areas. It achieves state-of-the-art performance on challenging KITTI, VKITTI2, and nuScenes datasets.
    % which strengthens the long-range temporal and context interaction with the context-aware temporal attention and long-range geometry embedding module.

    \item  We design a novel depth-pose context-aware temporal attention (CTA), which implicitly learns the temporal geometric constraints for moving objects via attention-based integration. 
    % It achieves the feature consistency and estimation integrity for moving objects, without additional explicit motion prediction or disentanglement module.

    \item We introduce a novel long-range geometry embedding (LGE) module to promote geometry reasoning among the long-range temporal frames.

    \item We develop an effective multi-level attention enhancement (MAE) module to make global-aware initial depth and estimations. It promotes the distinction of far-away small objects from the static background.
    % which injects cross-scale attention layers to enhance the interaction of multi-scale features, thus 
    
    % \item Our CTA-Depth pipeline achieves the state-of-the-art performance on challenging KITTI, VKITTI2, and nuScenes datasets.

\end{itemize}

\begin{figure*}[!t]  
        \centering
	\includegraphics[width =0.98 \textwidth]{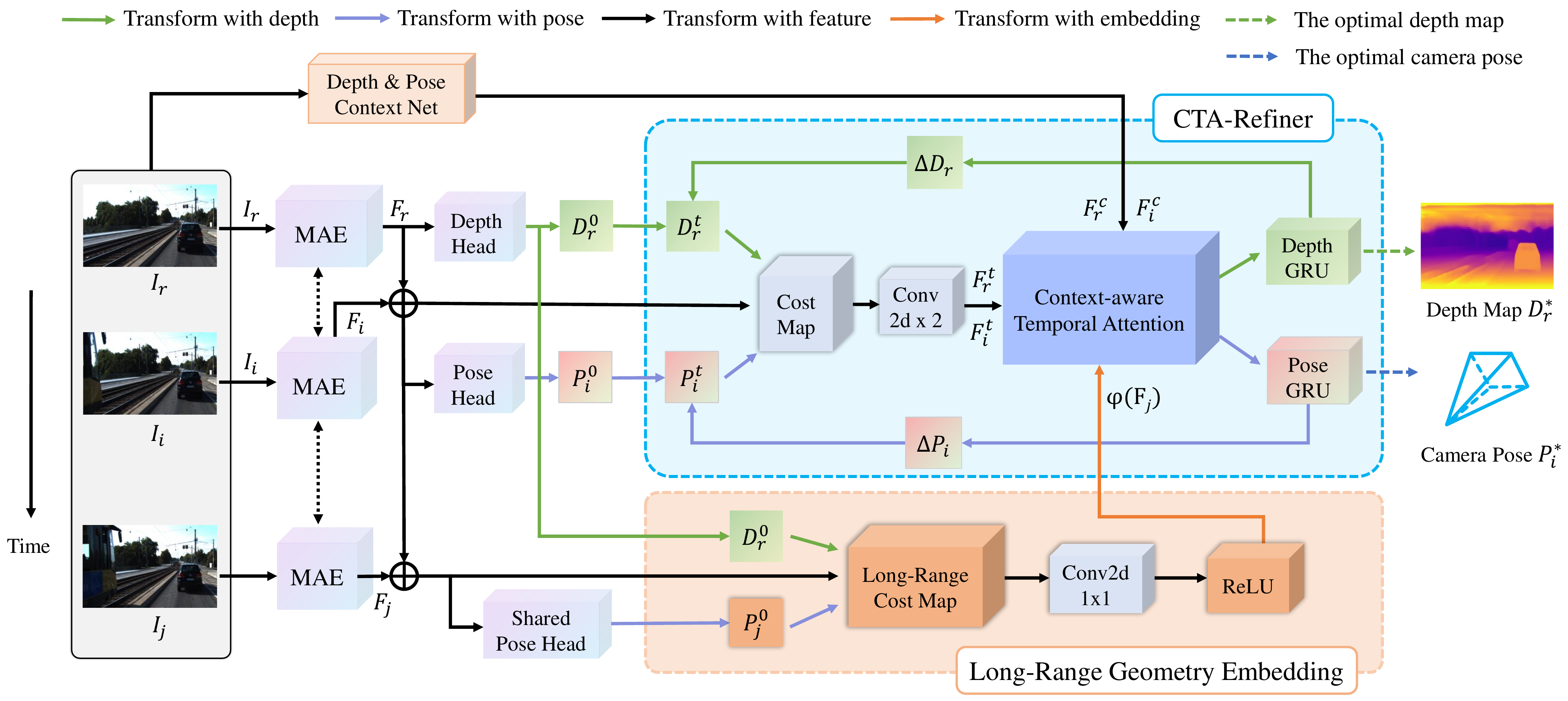}
	\caption{
		Overview of our CTA-Depth. Given a reference image $I_r$ and its $N$ monocular sequential images $\{I_i\}_{i=1}^N$, we first group $N$ pairs of network inputs, where each pair is composed of $I_r$ and a sequential image $I_i$. We then feed the two images into the MAE module and go through two heads to obtain the initial depth $D_r^0$ ,and pose $P_i^0$. Hence, the CTA-Refiner is proposed to alternately update the depth map and pose through iterations until the optimum solution, $D_r^*$ and $P_i^*$. In particular, with multiple temporal-neighboring frames $I_j$, we also design a long-range geometry embedding module to provide long-range temporal geometric priors for the depth refiner efficiently.
	}
	\vspace{-2mm}
    \label{fig:overview}
\end{figure*}

\section{Related Work}
\label{rel_work}

\paragraph{Monocular Depth Estimation.}
Convolutional neural networks with LiDAR supervision \cite{wang2015towards,fu2018deep,tang2018ba,teed2019deepv2d,lee2020multi,guizilini2021sparse,lee2021patch} have shown promising results in monocular depth estimation. As a pioneer, Eigen et al.~\cite{eigen2014depth} directly regressed depth by employing two stacked deep networks that made a coarse prediction from the whole image and then refined it locally.
On the other hand, \cite{laina2016deeper} adopted an end-to-end single CNN architecture with residual learning. 
To guide the encoded features to the desired depth prediction, \cite{lee2019big} further deployed it with local planar guidance layers. 
Recently, \cite{ranftl2021vision} introduced a dense prediction transformer ~\cite{Vaswani2017attention} for depth prediction. 
Meanwhile, \cite{bhat2021adabins,yang2021transformers} developed global information processing with vision transformer~\cite{dosovitskiy2020image} for performance boost. 
Besides, \cite{yuan2022newcrfs} adopted the swin-transformer \cite{liu2021swin} as the image encoder and the neural window fully-connected conditional random fields (NeWCRFs) module as the feature decoder. 
In particular, inspired by the RAFT \cite{teed2020raft} which employed a GRU-based recurrent operator to update optical flow, \cite{gu2023dro} proposed a multi-frame monocular approach with a deep recurrent optimizer to update the depth and camera poses alternately. However, these cost-map-based multi-frame methods \cite{teed2019deepv2d,gu2023dro} lead to performance degradation within dynamic areas due to the static scene assumption. To solve this problem, we introduce a long-range geometry embedding module and effectively inject the proposed depth-pose context-aware temporal attention into the deep refinement network for the optimization process of depth and pose.

\paragraph{Depth Estimation in Dynamic Environment}  Moving objects significantly hamper the multi-frame matching strategy due to the inevitable object truncation and occlusion.
Specifically, both the re-projection loss calculation and the cost volume construction fall into catastrophic failure cases during the positional change of observed targets. 
Existing works~\cite{li2019learning,lee2021learning,watson2021temporal} thus leveraged the segmentation mask to separate the static-scene depth loss from moving objects.
In particular, they also proposed explicit object motion prediction and a disentanglement module to assist the cost volume construction.
Specifically, SGDepth \cite{klingner2020self} proposed a semantic masking scheme providing guidance to prevent dynamic objects from contaminating the photometric loss. 
DynamicDepth \cite{feng2022disentangling} introduced an object motion disentanglement module that takes dynamic category segmentation masks as input to explicitly disentangle dynamic objects.
Considering time-crucial tasks such as autonomous driving, instead of adopting explicit redundant static-dynamic separation algorithms, we developed an efficient implicit modeling pipeline with our novel context-aware temporal attention module.
Besides, noting that previous works limited their methods to a few frames only due to an increase in computational cost, we developed a novel geometry embedding module to effectively encode semantic guidance from long-range time series. As a result, our pipeline can dynamically interact with the long-range semantic flow with the current-frame spatial geometry in a fully differentiable manner and is thus available for real industry implementation.

\section{Method}
\label{sec:method}

\subsection{Overview}

We demonstrate the framework of our approach in Fig.~\ref{fig:overview}, which mainly consists of five components: the network inputs, the multi-level attention enhancement (MAE) module, the depth \& pose context net, and the context-aware temporal attention refiner (CTA-Refiner) which includes the depth-pose context-aware temporal attention (CTA) module and the last long-range geometry embedding (LGE) module.

Given a reference image $I_r$ within one video captured via a monocular camera, and its $N$ video frames $\{I_i\}_{i=1}^N$, our goal is to predict the $I_r$'s accurate depth $D_r^*$ and the relative camera poses $\{P_i^*\}_{i=1}^N$ for sequence image $I_i$ with respect to $I_r$.
Specifically, we first regard the monocular video frames as network inputs. 
Then, we adopt the multi-level attention enhancement (MAE) module to extract representative visual features for the following depth head and (shared) pose head, which produces the initial depth $D_r^0$ and initial pose $P^0$.
Meanwhile, we employ the depth \& pose context net to extract context features $F_r^c$ for depth and $F_i^c$ for pose.
In addition, the long-range geometry embedding module seizes the multi-frame temporal geometry knowledge to create the long-range geometry embedding (LGE) $\varphi(F_j)$, which is further provided to the CTA to assist the refinement.
Afterwards, we adopt the context-aware temporal attention refiner (CTA-Refiner) to alternately update the predicted depth map and camera pose, which gradually converges to the optimal depth $D_r^*$ and pose $P_i^*$.

\begin{figure}[t]        
	\center{\includegraphics[width =0.48\textwidth]{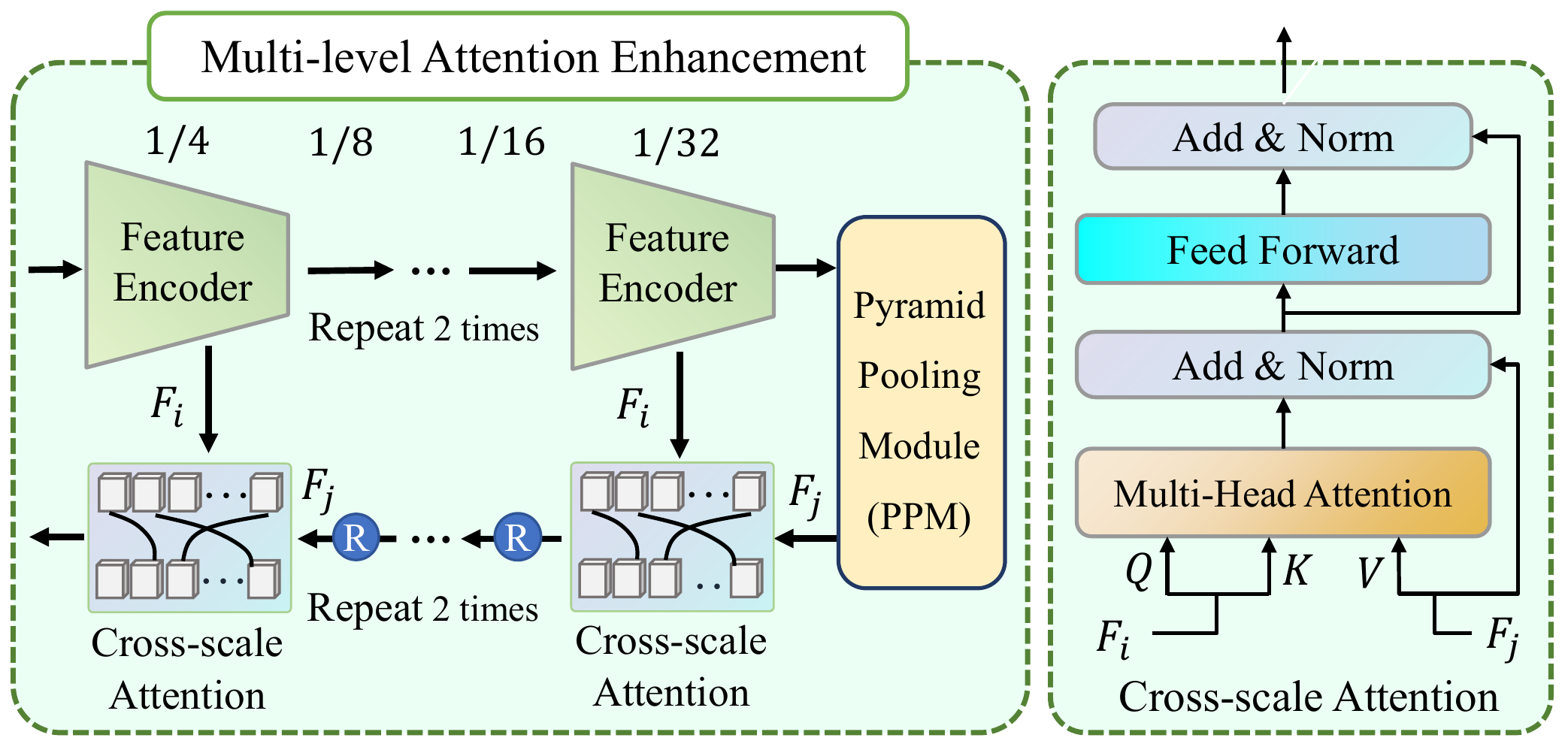}}
	\vspace{-2mm}
	\caption{The architecture of our multi-level attention enhancement (MAE) module, which adequately integrates multi-level image features via feature encoders, cross-scale attention layers, and PPM module. 
    ``$R$" denotes the rearranged up-scaling for feature maps.
    % Between each cross-attention layer, the rearrange upscale "$R$" is performed for feature map up-scaling.
	}
	\vspace{-2mm}
	\label{fig:encoder}
\end{figure}

\subsection{Multi-level Attention Enhancement (MAE)}
\label{subsec:embed}
Affiliated with the optimizer-based pipeline, we deliver our multi-level attention enhancement (MAE) module to achieve the initial prediction of depth and pose.
As shown in Fig.~\ref{fig:encoder}, we propose the multi-level feature setting and pyramid pooling module to reinforce the interest of the far-away small targets. 
In addition, to distinguish distant moving objects from the static background, we adopt cross-scale attention layers to enhance the interaction of different-scale features.
% Our proposed non-local cooperation 

Specifically, as shown in Fig.~\ref{fig:encoder}, we utilize the feature encoder to extract four different-scale features from $I_r$.
The low-level features focus on the local details, while the high-level features seize the global context information, which both contribute to the rising interest of the distant targets~\cite{lin2017feature,liu2018path}.
Afterwards, we employ the pyramid pooling module (PPM)~\cite{zhao2017pyramid} to aggregate these features and deliver four cross-scale attention layers to fuse the multi-level feature maps.
We use the scaled features $F_i$ as the query and key, and adopt the fusion features $F_j$ as the value to stimulate the interaction of multi-scale features. Within each cross-attention layer, we also introduce the rearranged up-scaling operation \cite{yuan2022newcrfs} to reduce the network complexity and boundary contour refinements. As shown in Fig.~\ref{fig005}, our method achieves accurate depth prediction of objects at different scales and long distances.
% Within each cross-attention layer, we also introduce the \gyz{rearrange up-scaling operation for feature up-scaling}, so that we could achieve robust context features for initial depth and pose prediction.  

% As above, but adding the rearrange upscale to reduce the parameters of the network. This module visually sharpens the boundary contours, though it does not achieve a large performance gain. Then the PPM is used to aggregate the non-local information of the whole image and consequently enhances the performance.

\begin{figure}[t]        
	\center{\includegraphics[width =0.48\textwidth]{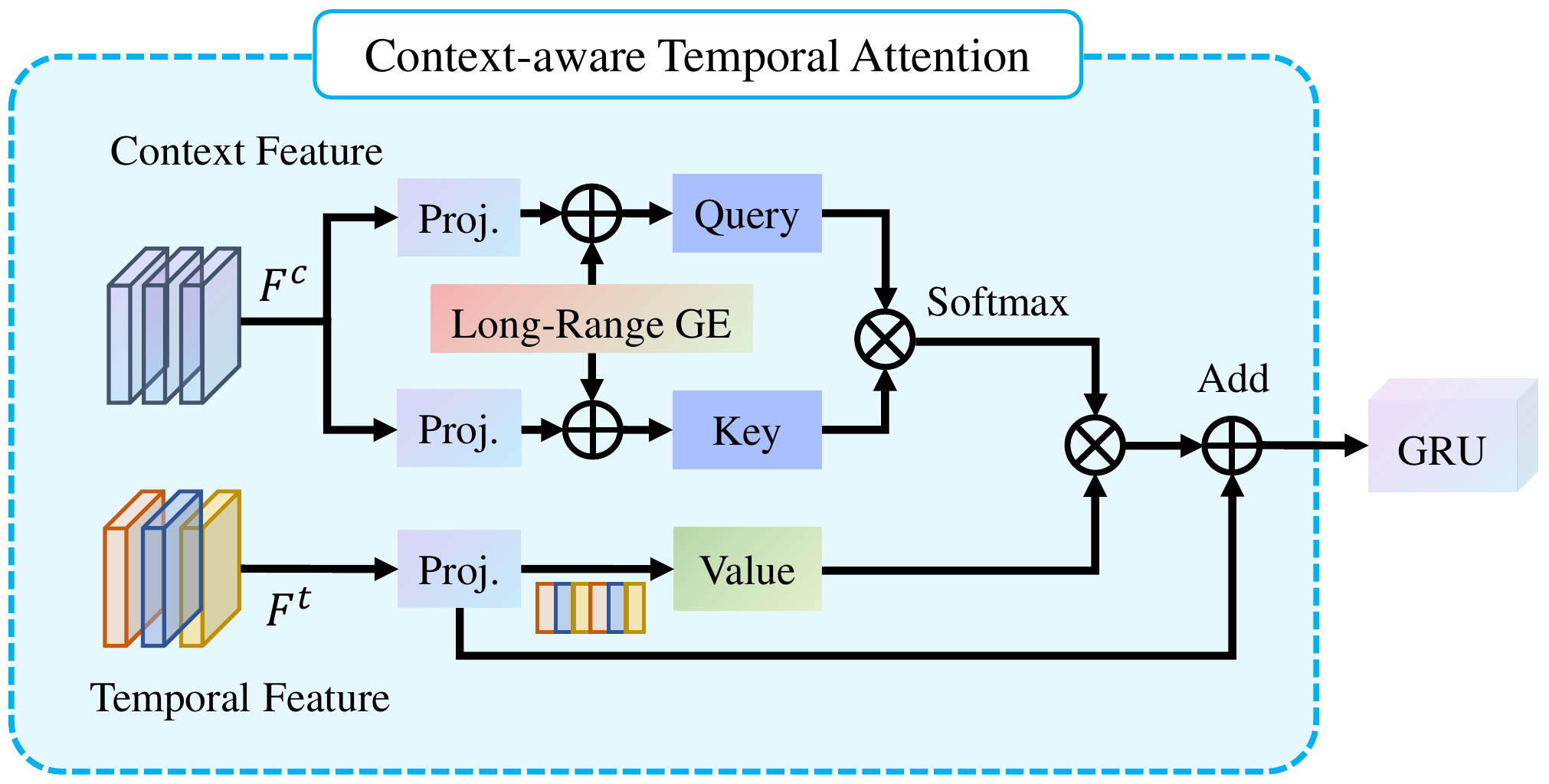}}
	\caption{
		Illustration of the Depth/Pose CTA. Our CTA-Refiner alternately optimizes the depth and the pose. During the depth refinement, the depth CTA uses the depth context feature $F_r^c$, the temporal feature $F_r^t$, and the Long-Range GE as inputs and feeds the outputs into the depth GRU. For the pose refinement, the pose CTA employs the pose context-aware feature $F_i^c$, the temporal feature $F_i^t$ and the Long-Range GE as inputs, then its outputs are fed into pose GRU.}
		\vspace{-2mm}
	\label{fig:optimizer}
\end{figure}

% \begin{figure}[t]        
% 	\center{\includegraphics[width =0.48\textwidth]{F4.pdf}}
% 	\caption{
% 		Illustration of the Depth/Pose CTA. Given the initial depth $D_r^0$ and relative camera pose $P_i^0$, CTA-Refiner alternately optimizes the depth or the pose. During the depth refinement (solid line), we first compute a cost map from the current solution $D_r^t$ and $P_i^t$, and extract depth temporal features $F_r^t$. Note that this $F_r^t$ contains the multi-frame temporal information from reference image $I_r$ and temporal sequential images $I_i$, since its inputs $D_r^t$ and $P_i^t$ are derived from $I_r$ and $I_i$. Then $F_r^t$ and the depth context features $F_r^c$ are fed into our depth context-aware temporal attention together. For the pose refinement (dashed line), we extract the pose temporal features $F_i^t$ and feed $F_i^t$ into the pose context-aware temporal attention along with the pose context features $F_i^c$.}
% 		\vspace{-2mm}
% 	\label{fig:optimizer}
% \end{figure}

\subsection{Context-aware Temporal Attention (CTA)}
\paragraph{CTA-Refiner}
We adopt the CTA-Refiner to iteratively refine initial estimations to the final converged depth $D_r^*$ and pose $P_i^*$, together with two introduced extra inputs: the context features $F_r^c$ for depth and $F_i^c$ for the pose from the depth \& pose context net, 
The refiner accepts these inputs to produce the prediction offset $\Delta D_r^t$ and $\Delta P_i^t$ and then updates the depth and pose as follows: 
\begin{align}
    D_r^{t+1} \leftarrow D_r^t + \Delta D_r^t, \\
    P_i^{t+1} \leftarrow P_i^t + \Delta P_i^t.
\end{align}
In detail, we first calculate the cost map given $D_r^t$, $P_i^t$, $F_r$ and $F_i$, as shown in Fig.~\ref{fig:overview}, where optimization for pose remains freezing when optimizing depth, and vice versa.
Notably, the cost map measures the distance in feature space between $I_r$ and the sequence image $I_i$.
Next, we adopt a simple feature extractor (two convolutions) to obtain the temporal features $F_r^t$ from the cost map, preparing for the following depth-pose CTA. 
Thus it implicitly rectifies the implied content-temporal inconsistency for moving objects and effectively promotes information integration between temporal features and depth-pose context features.

We formulate the cost map as the $L2$ distance between aligned feature maps $\mathcal{F}$. Given the depth map $D$ of the reference image
$I_r$ and the relative camera pose $T_i$ of another image $I_i$ with respect to $I_r$, the cost is constructed at each pixel $x$ in the reference image $I_r$:
\begin{equation}
	\mathbf{C}_{i}(x)=\left\|\mathcal{F}_{i}\left(\pi\left(\mathbf{T}_{i} \circ \pi^{-1}(x, \mathbf{D}(x))\right)\right)-\mathcal{F}_{r}(x)\right\|_{2},
\end{equation} 
where ${\pi()}$ is the projection of 3D points in 3D space onto the image plane and ${\pi^{-1}(x,{D}(x))}$ is the inverse projection. The transformation converts 3D points from the camera space of $I_r$ to that of $I_i$. The multiple cost values are averaged at each pixel ${x}$ for multiple neighboring images.

\paragraph{Depth-pose Context-aware Temporal Attention (CTA)} 
The construction of the cost volume heavily relies on the static scene assumption, where it supposes that the object points remain static at time $t$ and $t^*$.
Thus, we re-project the features at time $t$ to another plane with pose $t^*$ at time $t^*$, to achieve the matching cost values.
However, moving objects break this assumption since targets such as cars, trains, or pedestrians with a certain speed could move within the time gap.
This gives rise to the feature inconsistency deviation, degraded (mismatching) cost values and re-projection loss, and finally drawbacks our optimization for depth and pose. 
We discard explicit settings such as the object motion prediction module or disentangle module~\cite{lee2021learning,feng2022disentangling,wimbauer2021monorec}, which brings additional complexity and ignores the potential of complementary context-temporal information.
Instead, we deliver our context-aware temporal attention (CTA) to implicitly rectify the mismatching problem, which efficiently cooperates the temporal features with context features via attention-based communication to achieve the feature consistency and estimation integrity.

%-----------------------------
Specifically, as shown in Fig.~\ref{fig:optimizer}, taking depth optimization as 
an example, we first lift the temporal feature $F_r^t$ to value ($V$) vectors via the mapping function $\sigma(\cdot)$.
Meanwhile, we create query ($Q$) and key ($K$) vectors by adding the mapping functions $\theta(\cdot)$ and $\phi(\cdot)$ from the context feature $F_r^c$, and prepare long-range geometry embedding (LGE). 
We first allocate the query, key and value as  $Q = \theta(F_r^c)\oplus LGE, \  K = \phi(F_r^c)\oplus LGE$, and $V= \sigma(F_r^t)$, respectively.
Subsequently, the depth context-aware temporal attention is denoted as follows:
\begin{align}
F_r^d  = f_{\text{s}}(Q\otimes K)\otimes V \oplus F_r^t,
\end{align}

% \begin{equation}
% \label{equ:attention1}
% F_r^d = f_{\text{s}}((\theta(F_r^c)\oplus LGE)\otimes(\phi(F_r^c) \oplus LGE))\otimes\sigma(F_r^t) \oplus F_r^t,
% \end{equation}
where $f_s$ denotes the softmax operation, $\oplus$ denotes the point-wise addition and $\otimes$ denotes matrix multiplication. 

Intuitively, compared with directly feeding $F_r^t$ and $F_r^c$ for refinement, our CTA explicitly aligns the features for moving objects through the cross-attention mechanism, to compensate for the mismatching discrepancy, which guarantees temporal-context feature fusion and seamless depth refinement.
The temporal feature also helps the context feature to fulfill the moving target’s integrity, such as the second row of Fig.~\ref{fig:teaser}, where we rectify the wrong estimation in the ‘hole’ of the car head with temporal-content interaction.
Similarly, for the pose optimization with fixed depth optimization, we employ context features $F_i^c$ to extract query ($Q$) and key ($K$) vectors with added LGE.
Hence, we allocate the query, key and value as $Q = \theta(F_i^c)\oplus LGE, \   K = \phi(F_i^c)\oplus LGE$, and $ V= \sigma(F_i^t)$, respectively.
Particularly, we lift the temporal feature $F_i^t$ for value ($V$) vectors and finally adopt similar attention for pose context-aware temporal attention:
\begin{align}
F_i^p  = f_{\text{s}}(Q\otimes K)\otimes V \oplus F_i^t.
\end{align}
% \textcolor{red}{TOO LONG, SEPARATE THE FUNCTION INTO TWO}
% \begin{equation}
% \label{equ:attention2}
% F_i^d = f_{\text{s}}((\theta(F_i^c) \oplus LGE )\otimes (\phi(F_i^c) \oplus LGE ))\otimes\sigma(F_i^t) \oplus F_i^t.
% \end{equation}
%-------------------------
\subsection{Long-range Geometry Embedding (LGE)}
Existing works~\cite{gu2023dro,feng2022disentangling} usually adopt the two-frame correlation, such as the cost volume constructed from two frames, which limits the temporal interest field and wastes temporal helpful clues for geometry reasoning within the originally long-range multiple frames.
Therefore, to expand the temporal geometry relation range, especially for moving targets within long-range frames, we present our long-range geometry embedding module to create long-range geometry embedding (LGE), which seizes the beneficial long-range temporal hints for geometry and is inserted into the depth-pose CTA to promote the present depth and pose refinement.

Specifically, we demonstrate the detailed structure in Fig.~\ref{fig:overview}, where we build the long-range cost map with the reference frame $I_r$ and another arbitrary frame $I_j (j\neq{i})$.
Hence, we can achieve N-1 cost maps with N-1 different-range frames.
Notably, we achieve the temporal feature with long-range cost maps and convolution, which reveals the same as the procedure in CTA-Refiner. 
Afterwards, we propose the $1\times1$ Conv, ReLU and Conv to deliver the geometry embedding $\varphi (F_j)$.
Specifically, the number of $\varphi (F_j), j\neq{i}$ is N-1.
Then, we aggregate this long-range temporal knowledge to create our embedding $F_{ge}$ via the addition operation:
\begin{equation}
    F_{ge}=\sum_{j, j\neq i}^{N} \varphi (F_j),
\end{equation}
which will be added to the query and key vectors within the depth-pose CTA.
Our approach efficiently reveals most of the geometry clues within the long-range temporal frames.
It is also regarded as a temporal geometry prior to enhance the temporal-context association process.

\subsection{Supervised Training Loss}
We train our network by optimizing both depth and pose errors.
Then, we formulate the depth loss as the $L1$ distance between the predicted depth map $D$ and the associated ground truth $\hat{D}$:
\begin{equation}
	\mathcal{L}_{\rm depth} = \sum_{s=1}^m \gamma^{m-s} \Arrowvert { {D}}^s - \hat{ {D}} \Arrowvert_1,
\end{equation}
where the discounting factor $\gamma$ is 0.85 and $s$ denotes the stage number. 
There are $m$ alternating update stages of depth and pose refinements. At each stage, we repeatedly refine the depth and pose $n$ times.
%during recurrent optimization.
Next, our pose loss is defined based on the ground truth depth $\hat{D}$ and pose ${\rm {\hat T}_i}$ with $I_i$ relative to the reference image $I_r$: 
\begin{equation}
\begin{aligned}
	\mathcal{L}_{\rm pose} = &\sum_{s=1}^m\sum_{x} \gamma^{m-s} \Arrowvert \pi({\rm {T}}_i^s \circ \pi^{-1}(x, \hat{D}(x)))\\
	&- \pi({\hat{\rm {T}}}_i \circ \pi^{-1}(x,\hat{D}(x)))  \Arrowvert_1 ,\\
	\end{aligned}
\end{equation}
where $\circ$ means the Hadamard product. 
The pose loss summarizes the re-projection deviation of the pixel $x$ according to the estimated camera pose ${\rm {T}}_i^s$ and the true pose ${\hat{\rm {T}}}_i$ in each stage. 
${\pi()}$ is the projection of 3D points in 3D space onto the image plane. Its inverse projection ${\pi^{-1}(x,\hat{D}(x))}$ maps the pixel ${x}$ and its ground truth depth $\hat{D}(x)$ back points in the 3D space. 
Finally, the total supervised loss is calculated by:
\begin{equation}
	\mathcal{L}_{\rm supervised} = \mathcal{L}_{\rm depth} + \mathcal{L}_{\rm pose} . 
\end{equation}

\section{Experiments and Results}
\label{sec:experiments}

\begin{table*}[t]
	\centering
% 	Note that, "M→O" means monocular multiple frame images are used in training while only single frame image is used for inference. and "MF" means monocular multi-frames are both used in traning and inference.“SF” means single frame.
	\resizebox{\textwidth}{!}{%
		\begin{tabular}{@{}lccccccccccc@{}}
			\toprule
			Method            & Cap    & Input          & GT type & Abs Rel $\downarrow$  & Sq Rel $\downarrow$ & RMSE $\downarrow$ & RMSE$_{log}$  $\downarrow$   & $\delta_1 < 1.25$ $\uparrow$ & $\delta_2 < 1.25^2$ $\uparrow$ & $\delta_3 < 1.25^3$ $\uparrow$ \\ \midrule
			PackNet-SfM \cite{guizilini20203d}       & 0-80m     & M$\to${S}         &Velodyne& 0.090& 0.618& 4.220& 0.179& 0.893& 0.962& 0.983         \\
			DRO \cite{gu2023dro}                & 0-80m       & Multi-Frame                &Velodyne& 0.073& 0.528& 3.888& 0.163& 0.924& 0.969& 0.984         \\
			CTA-Depth (ours)            &         0-80m      & Multi-Frame               & Velodyne         &   0.071      &     0.496   &   3.598    &   0.143    &   0.931    &    0.975     &  0.989      \\
			\midrule
			BTS \cite{lee2019big}              & 0-80m     & Single-Frame         & Improved & 0.059 & 0.241  & 2.756 & 0.096 & 0.956 & 0.993   & 0.998   \\
			GLPDepth \cite{kim2022global}    & 0-80m        & Single-Frame                &Improved& 0.057& \multicolumn{1}{c}{--}& 2.297& 0.086& 0.967& 0.996& 0.999 \\
			PackNet-SfM \cite{guizilini20203d}       & 0-80m    & M$\to${S}                & Improved & 0.064 & 0.300  & 3.089 & 0.108 & 0.943 & 0.989   & 0.997   \\
			\midrule
			BANet \cite{tang2018ba}   & 0-80m & Multi-Frame & Improved & 0.083 & \multicolumn{1}{c}{--} & 3.640 & 0.134 & \multicolumn{1}{c}{--} & \multicolumn{1}{c}{--} & \multicolumn{1}{c}{--}  \\
			DeepV2D(2-view) \cite{teed2019deepv2d}    & 0-80m    & Multi-Frame                    &Improved& 0.064& 0.350& 2.946& 0.120& 0.946& 0.982& 0.991         \\
			DRO \cite{gu2023dro}         & 0-80m     & Multi-Frame                   &Improved& 0.047& 0.199& 2.629& 0.082& 0.970& 0.994& 0.998        \\
                \textbf{CTA-Depth}            &         0-80m        & Multi-Frame             &   Improved       & \textbf{0.038}       &    \textbf{0.145}  &   \textbf{2.224}    &  \textbf{0.069}     &   \textbf{0.978}    &    \textbf{0.996}     & \textbf{0.999}      \\
			% \textbf{CTA-Depth}            &         0-80m        & Multi-Frame             &   Improved       & 0.038       &    0.157  &   2.288    &  0.071     &   0.977    &    \textbf{0.996}     & \textbf{0.999}      \\
			% \textbf{CTA-Depth(L)}            &         0-80m        & Multi-Frame             &   Improved       &  \textbf{0.037}       &     \textbf{0.139}   &   \textbf{2.191}    &  \textbf{0.068}     &   \textbf{0.980}    &    \textbf{0.996}     &   \textbf{0.999}      \\
			% \midrule
			% % ResNet-18 + TransOpt,Swin-L + TransOpt	
   %              \textit{Improvement v.s. Multi-Frame Baseline} & \multicolumn{1}{c}{--} & \multicolumn{1}{c}{--}& \multicolumn{1}{c}{--} & \textcolor{blue}{19.1$\%$} $\uparrow$ &  \textcolor{blue} {27.1$\%$} $\uparrow$ &  \textcolor{blue}{15.4$\%$} $\uparrow$ &  \textcolor{blue}{15.8$\%$} $\uparrow$ &  \textcolor{blue}{0.8$\%$} $\uparrow$ &  \textcolor{blue}{0.2$\%$} $\uparrow$& \multicolumn{1}{c}{--}\\
			% \textit{Improvement v.s. Multi-Frame Baseline} & \multicolumn{1}{c}{--} & \multicolumn{1}{c}{--}& \multicolumn{1}{c}{--} & \textcolor{blue}{21.3$\%$} $\uparrow$ &  \textcolor{blue} {30.2$\%$} $\uparrow$ &  \textcolor{blue}{16.7$\%$} $\uparrow$ &  \textcolor{blue}{17.1$\%$} $\uparrow$ &  \textcolor{blue}{1.0$\%$} $\uparrow$ &  \textcolor{blue}{0.2$\%$} $\uparrow$& \multicolumn{1}{c}{--}\\
			\bottomrule
		\end{tabular}%
    }
    \caption{Quantitative results of supervised monocular depth estimation methods on the KITTI Eigen split. Note that the seven widely-used metrics are calculated strictly following the baseline \protect\cite{gu2023dro} and ground-truth median scaling is applied. "M$\to$S" means monocular multiple frame images are used in training while only a single frame image is used for inference. We utilize bold to highlight the best results.}
    \label{table_kitti_1}
    \vspace*{-2mm}
\end{table*}

\begin{table*}[t]
    \centering
	\footnotesize
	\setlength{\tabcolsep}{3.5pt}
	% For LaTeX tables use
% 	\vspace{-0.5em}
	\centering
	\resizebox{\textwidth}{!}
	{%
		\begin{tabular}{lcccccccccc}
			\toprule
			Method & Reference & Input & Abs Rel $\downarrow$  & Sq Rel $\downarrow$ & RMSE $\downarrow$ & RMSE$_{log}$  $\downarrow$ & $\delta_1 < 1.25$ $\uparrow$ & $\delta_2 < 1.25^2$ $\uparrow$ & $\delta_3 < 1.25^3$ $\uparrow$ & FPS $\uparrow$\\  
			\midrule
			Xu et al. \cite{xu2018structured} & CVPR 2018 & Single-Frame & 0.122 & 0.897 & 4.677 & \multicolumn{1}{c}{--} &  0.818 &  0.954 & 0.985 & \multicolumn{1}{c}{--} \\
			DORN \cite{fu2018deep} & CVPR 2018 & Single-Frame & 0.072 & 0.307 & 2.727 & 0.120 & 0.932 & 0.984 & 0.995 & \multicolumn{1}{c}{--} \\
			Yin et al. \cite{yin2019enforcing} & ICCV 2019 & Single-Frame & 0.072 & \multicolumn{1}{c}{--} &  3.258 &  0.117 &  0.938 &  0.990 & 0.998 & \multicolumn{1}{c}{--} \\
			PackNet-SAN \cite{guizilini2021sparse} & CVPR 2021 & Single-Frame & 0.062 & \multicolumn{1}{c}{--} &  2.888 &  \multicolumn{1}{c}{--} &  0.955 &  \multicolumn{1}{c}{--} & \multicolumn{1}{c}{--} & \multicolumn{1}{c}{--} \\
			DPT* \cite{ranftl2021vision} & ICCV 2021 & Single-Frame & 0.062 & \multicolumn{1}{c}{--} &  2.573 &  0.092 &  0.959 &
			0.995 & 0.999 & \multicolumn{1}{c}{--} \\
			PWA \cite{lee2021patch} & AAAI 2021 & Single-Frame & 0.060 & 0.221 &  2.604 &  0.093 &  0.958 &  0.994 & 0.999 & \multicolumn{1}{c}{--} \\
			AdaBins \cite{bhat2021adabins} & CVPR 2021 & Single-Frame & 0.058 & 0.190 &  2.360 &  0.088 & 0.964 & 0.995 & 0.999 & 2.96 \\
			NeWCRFs \cite{yuan2022newcrfs} & CVPR 2022 & Single-Frame & \textbf{0.052} & \textbf{0.155} &  \textbf{2.129} &  \textbf{0.079} &  \textbf{0.974} &  \textbf{0.997} & \textbf{0.999} & \textbf{3.48} \\
			P3Depth \cite{patil2022p3depth} & CVPR 2022 & Single-Frame & 0.071 & 0.270 &  2.842 &  0.103 &  0.953 &  0.993 & 0.998 & \multicolumn{1}{c}{--} \\
% 			\midrule
% 			PackNet-SfM \cite{guizilini20203d}       & CVPR 2020    & M$\to${O} & 0.064 & 0.300  & 3.089 & 0.108 & 0.943 & 0.989   & 0.997 & \multicolumn{1}{c}{--} \\
			\midrule
			BANet \cite{tang2018ba}   & ICLR 2019  & Multi-Frame & 0.083 & \multicolumn{1}{c}{--} & 3.640 & 0.134 & \multicolumn{1}{c}{--} & \multicolumn{1}{c}{--} & \multicolumn{1}{c}{--} & \multicolumn{1}{c}{--}\\
% 			BA-Net9.51
			DeepV2D \cite{teed2019deepv2d}   & ICLR 2020  & Multi-Frame & 0.064& 0.350& 2.946& 0.120& 0.946& 0.982& 0.991 & 0.67 \\
            MaGNet \cite{bae2022multi} & CVPR 2022 & Multi-Frame & 0.054 & 0.162 & \textbf{2.158} &  0.083 &  0.971 &  \multicolumn{1}{c}{--} & \multicolumn{1}{c}{--} & \multicolumn{1}{c}{--} \\
            DRO \cite{gu2023dro} & RA-L 2023 & Multi-Frame & 0.059 & 0.230 &  2.799 &  0.092 &  0.964 &  0.994 & 0.998 & \textbf{6.25} \\
            % MonoRec & \multicolumn{1}{c}{--} & Multi-Frame & 0.050 & 0.295 &  2.266 &  0.082 &  0.973 &  0.991 & 0.9986 & \multicolumn{1}{c}{--} \\
         \textbf{CTA-Depth} & \multicolumn{1}{c}{--} & Multi-Frame &  \textbf{0.045} & \textbf{0.156} &  2.275 & \textbf{0.073} &  \textbf{0.978} &  \textbf{0.997} & \textbf{0.999} & 5.53\\
			\bottomrule		
		\end{tabular}%
	}
 \caption{Quantitative results on KITTI Eigen split with the cap of 0-80m. 
 Note that the seven widely-used metrics are calculated  strictly following AdaBins  \protect\cite{bhat2021adabins}. 
 ``Abs Rel" error occupies the main ranking metric. 
% ``FPS" of some recent methods is calculated on the same Nvidia RTX A6000 GPU. 
 ``*" means using additional data for training. 
 We utilize bold to highlight the best results of single-frame methods and multi-frame methods.
 }
\label{table_kitti_2}       % Give a unique label
\vspace{-4mm}
\end{table*}
\subsection{Datasets}
We evaluate our method on three public benchmark datasets, including KITTI, Virtual KITTI 2, and nuScenes. These datasets provide a large number of monocular temporal neighboring images in dynamic environments.
%Below, we shall briefly introduce each one of them.
% \vspace{-2mm}
%, including KITTI, Virtual KITTI 2, and nuScenesand the preferred training dataset both in the supervised and self-supervised monocular depth estimation. It contains over 93 thousand depth maps with corresponding raw LiDAR scans and RGB images, aligned with the raw data. Our work performs monocular depth estimation on the widely used KITTI Eigen split \cite{eigen2014depth}. This split is composed of 22,600 images from 32 scenes for training and 697 images from 29 scenes for testing with LiDAR-based ground truth. The corresponding depth of each RGB image is sampled sparsely by the rotating LiDAR sensor.

\noindent\textbf{KITTI} \cite{geiger2012we} is a popular benchmark for the task of autonomous driving, which provides over 93,000 depth maps with corresponding raw LiDAR scans and RGB images aligned with raw data. In experiments, we follow the widely-used KITTI Eigen split \cite{eigen2014depth} for network training, which is composed of 22,600 images from 32 scenes for training and 697 images from 29 scenes for testing. The corresponding depth of each RGB image is sampled sparsely by the LiDAR sensor.

\noindent\textbf{Virtual KITTI 2} \cite{gaidon2016virtual} is widely used for video understanding tasks, which consists of 5 sequence clones from the KITTI tracking benchmark and contains 50 high-resolution monocular videos generated from five different virtual worlds in urban settings under various imaging and weather conditions. These photo-realistic synthetic videos are fully annotated with depth labels.

\noindent\textbf{nuScenes} \cite{caesar2020nuscenes} is a large-scale multi-modal autonomous driving dataset that is the first to carry the completely autonomous vehicle sensor suite: 32-beam LiDAR, 6 cameras and 5 radars with 360$^\circ$ coverage. It comprises 1,000 scenes, where each scene lasts 20 seconds and is fully annotated with 3D bounding boxes for 23 classes and 8 attributes.

\subsection{Implementation Details}
We implement our CTA-Depth in PyTorch and train it for 100 epochs with a mini-batch size of 4. 
The learning rate is $2\times10^{- 4}$ for both depth and pose refinement, which is decayed by a constant step (gamma=$0.5$ and step size=$30$). 
We set $\beta_1=0.9$ and $\beta_2=0.999$ in the Adam optimizer. 
We resize the input images to $320\times960$ for training, and set the number of sequential images to 2 for CTA-Refiner by balancing both computation efficiency and prediction accuracy. For long-range geometry embedding, the number of temporally adjacent images is set to $N=3$. Since the output $\varphi(F_j)$ of the LGE for the same image is fixed and not updated with the iterations, this provides more prior temporal information to CTA-Refiner while ensuring network efficiency.
% \textcolor{red}{ADD NUMBER OF FRAMES USED FOR EMBEDDING}.
We fix $m$ at 3 and $n$ at 4 in experiments.

% set the number to 2, not set the number to be 2
% As described in Section \nameref{subsec:embed}, we have two solutions: CTA-Depth and CTA-Depth (L).
%We trained our network on four Nvidia RTX A6000 GPUs. 
%Generally speaking, it takes about two days for training on KITTI dataset.
% Generally speaking, it takes about two days for training and \xz{$XX$} hours for inference on KITTI dataset.

%deployed it on \xz{XX} Nvidia RTX A6000 GPUs. The network is trained for 200 epochs with a batch size of 4. For the optimizer, we employ Adam and set $\beta_{\text{1}}$ to 0.9 and $\beta_\text{2}$ to 0.999. The learning rate is set to 0.0002 for both depth and pose head and is decayed by constant step (gamma = 0.5, step size = 30). The input image is resized as 320 * 960. We use the loss functions and two-view setting proposed by previous excellent work \cite{gu2023dro} to optimize the TransOpt network. The training losses for supervised and self-supervised schemes are used for our monocular depth estimation tasks. In detail, the $\gamma$ is set to 0.85 in supervised training. The $\alpha$ is set to 0.85 and the $\lambda$ is set to 0.01 in self-supervised training.
\begin{figure*}[ht]
	\centering
	\begin{minipage}{0.32\linewidth}
        \centerline{\includegraphics[width=\textwidth]{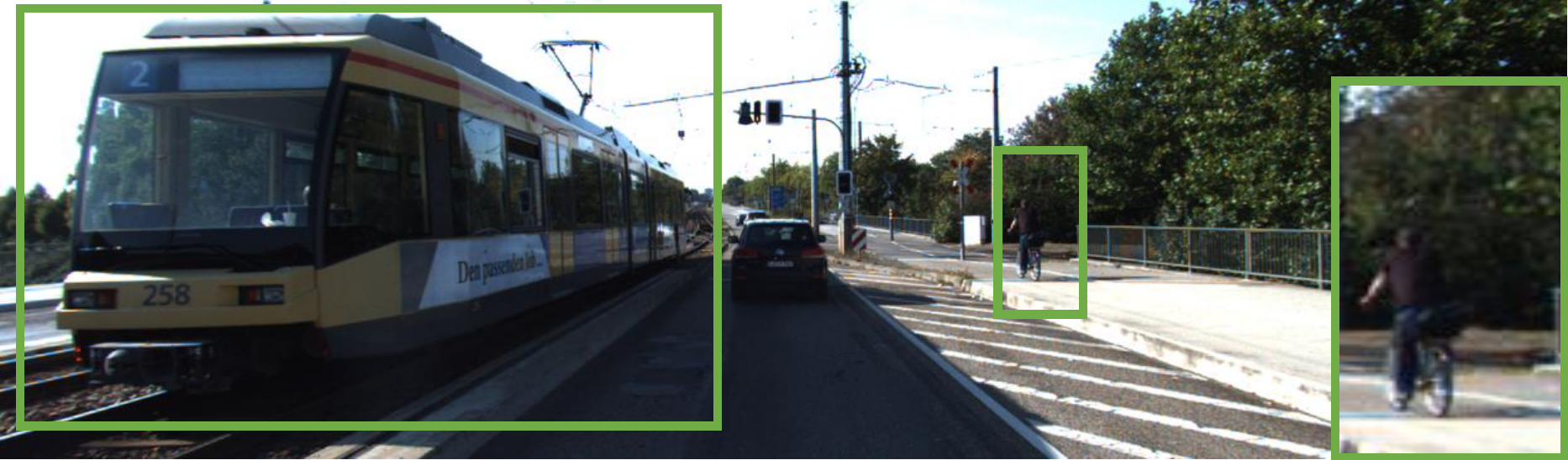}}
		\vspace{0.8pt}
% 		\centerline{\includegraphics[width=\textwidth]{Images/demo_img_1/0000000140_1.png}}
% 		\vspace{0.8pt}
		\centerline{\includegraphics[width=\textwidth]{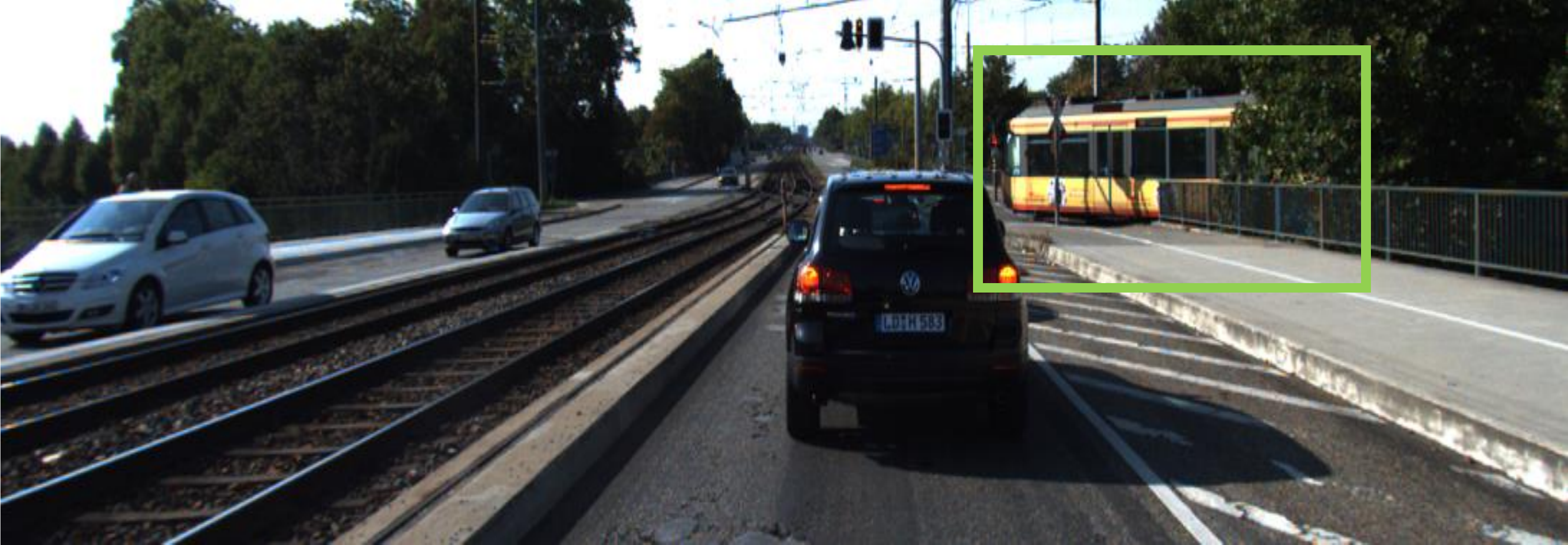}}
		\vspace{0.8pt}
		\centerline{\includegraphics[width=\textwidth]{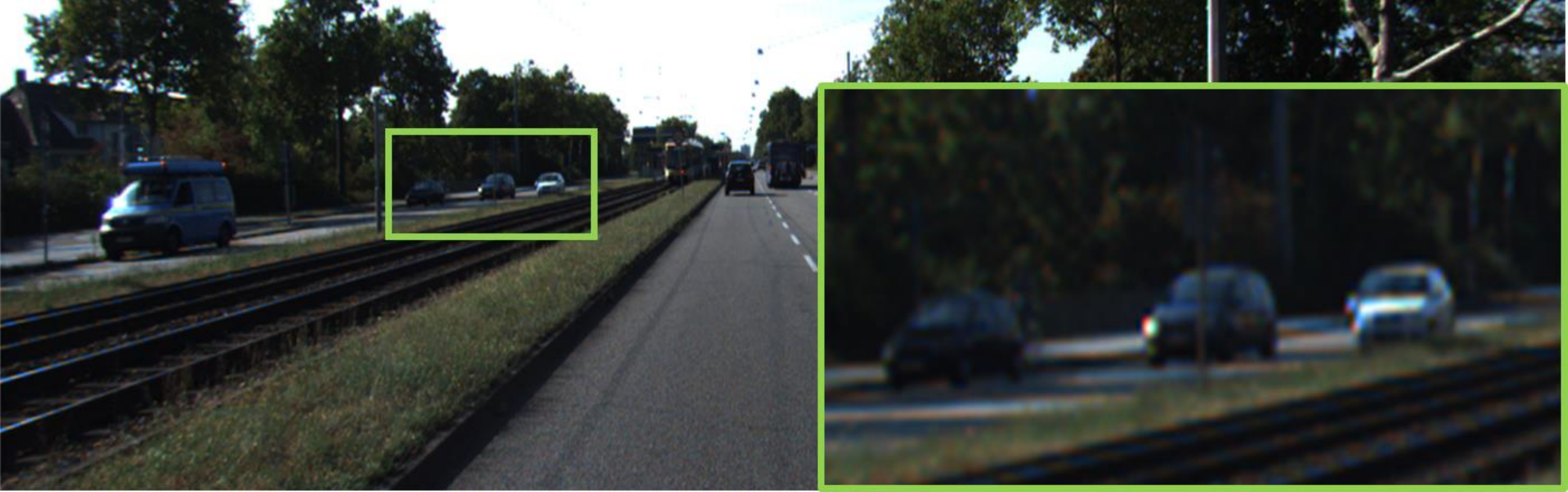}}
		\vspace{0.8pt}
% 		\centerline{\includegraphics[width=\textwidth]{Images/demo_img_1/0000000153_1.png}}
% 		\vspace{0.8pt}	
		\begin{footnotesize}
			\centerline{(a) Input images}
		\end{footnotesize}
	\end{minipage}
	\begin{minipage}{0.32\linewidth}
		%\centerline{\includegraphics[width=\textwidth]{Images/demo_img_2/0000000021_1.png}}
		%\vspace{0.8pt}
% 		\centerline{\includegraphics[width=\textwidth]{Images/demo_img_2/0000000095_1.png}}
% 		\vspace{0.8pt}
% 		\centerline{\includegraphics[width=\textwidth]{Images/demo_img_2/0000000113_1.png}}
		\centerline{\includegraphics[width=\textwidth]{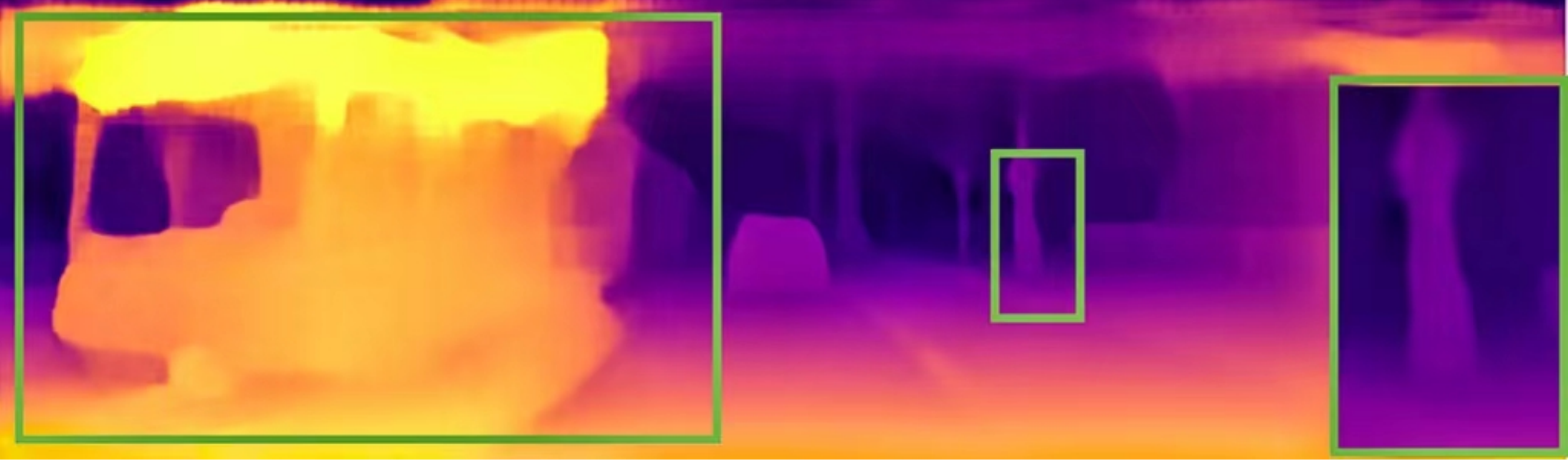}}
		\vspace{0.8pt}
		\centerline{\includegraphics[width=\textwidth]{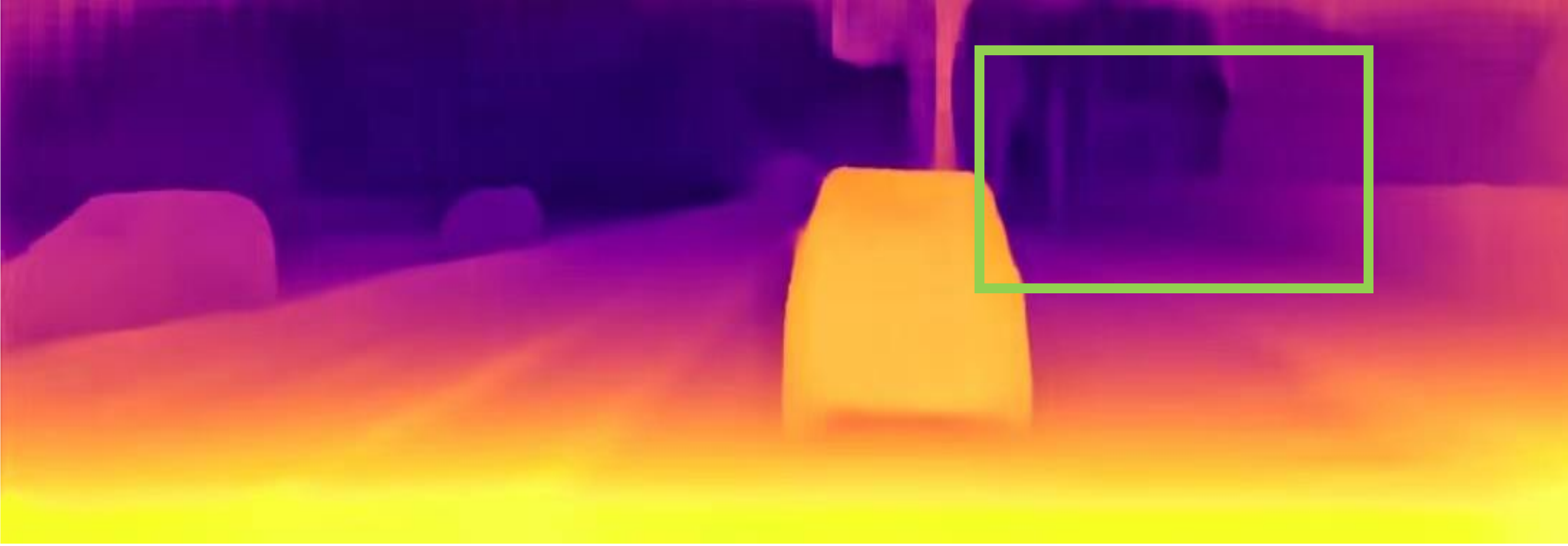}}
		\vspace{0.8pt}
		\centerline{\includegraphics[width=\textwidth]{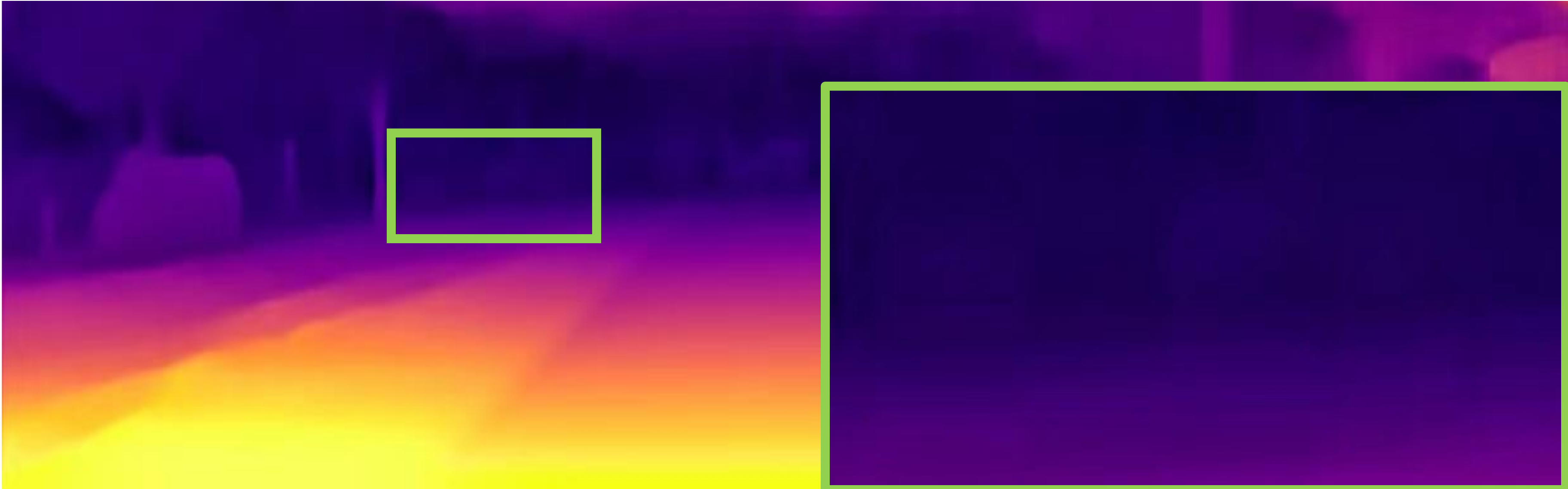}}
		\vspace{0.8pt}
% 		\centerline{\includegraphics[width=\textwidth]{Images/demo_img_2/0000000153_1.png}}
% 		\vspace{0.8pt}	
		\begin{footnotesize}
			\centerline{(b) Results of the Baseline}
		\end{footnotesize}
	\end{minipage}
	\begin{minipage}{0.32\linewidth}
		%\centerline{\includegraphics[width=\textwidth]{Images/demo_img_3/0000000021_1.png}}
		%\vspace{0.8pt}
% 		\centerline{\includegraphics[width=\textwidth]{Images/demo_img_3/0000000095_1.png}}
% 		\vspace{0.8pt}
% 		\centerline{\includegraphics[width=\textwidth]{Images/demo_img_3/0000000113_1.png}}
		\centerline{\includegraphics[width=\textwidth]{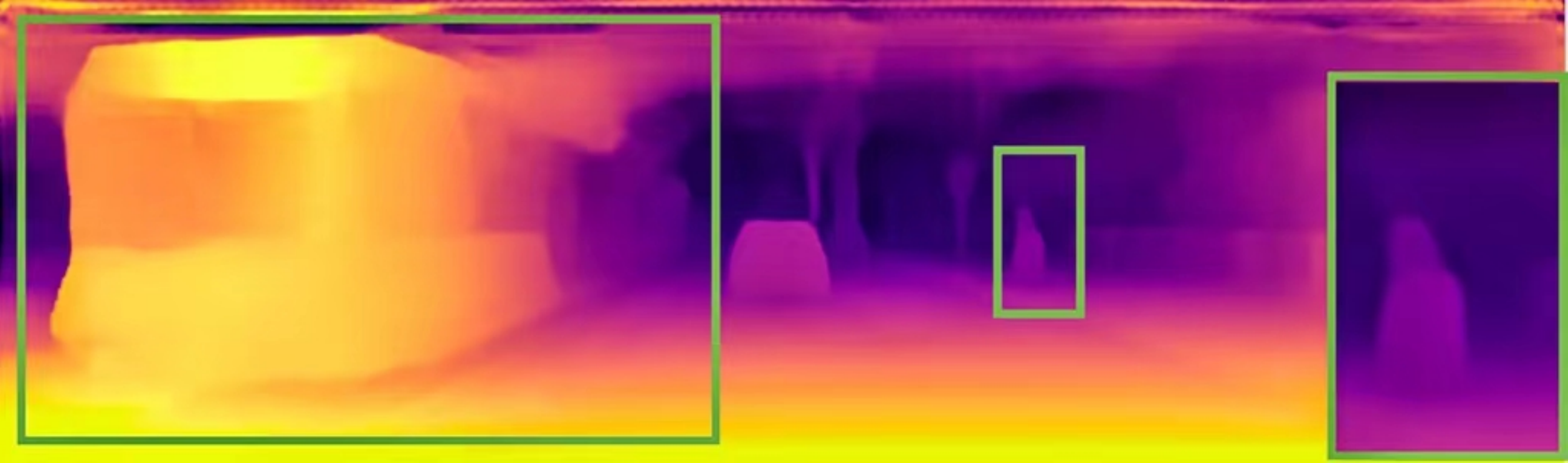}}
		\vspace{0.8pt}
% 		\centerline{\includegraphics[width=\textwidth]{Images/demo_img_3/0000000140_1.png}}
% 		\vspace{0.8pt}
		\centerline{\includegraphics[width=\textwidth]{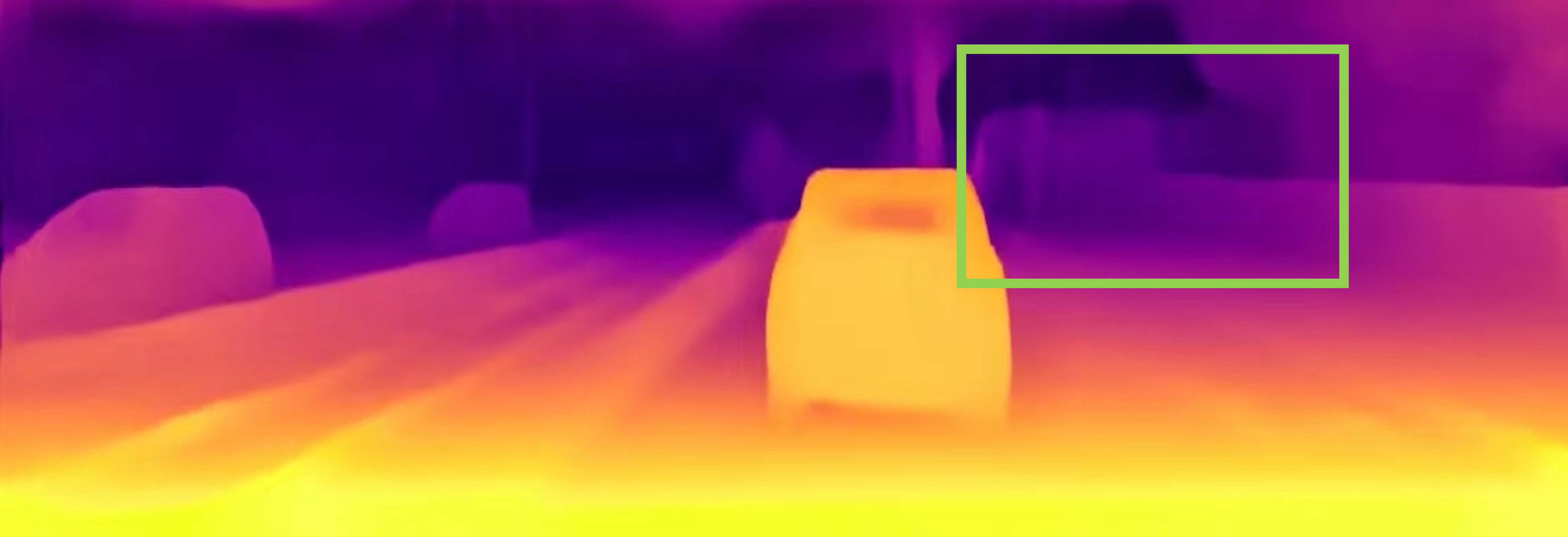}}
		\vspace{0.8pt}
		\centerline{\includegraphics[width=\textwidth]{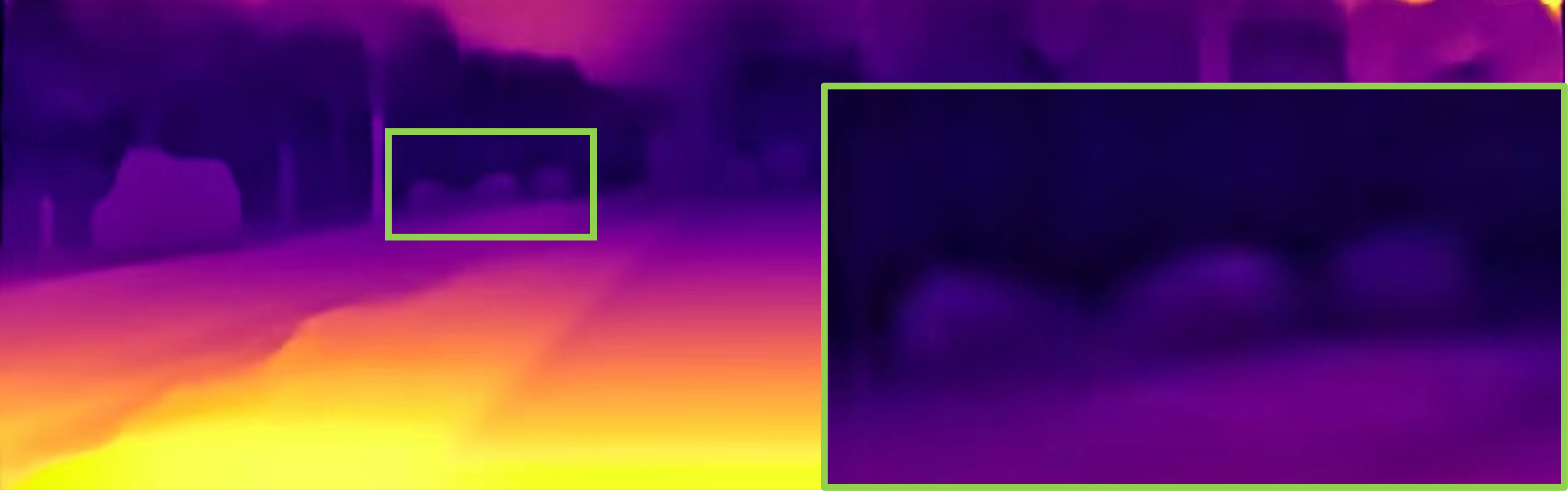}}
		\vspace{0.8pt}
% 		\centerline{\includegraphics[width=\textwidth]{Images/demo_img_3/0000000153_1.png}}
% 		\vspace{0.8pt}
		\begin{footnotesize}
			\centerline{(c) CTA-Depth (Ours)}
		\end{footnotesize}
	\end{minipage}
	\caption{Qualitative comparisons given input images (a) from KITTI. Clearly, our method (c) yields more accurate depth maps than the Baseline method; (b) see particularly the regions marked by green boxes.}
	%Green rectangles indicate the important regions for result analysis. }
	\vspace*{-2mm}
	\label{fig005}
\end{figure*}
\subsection{Computation Time Analysis}
Given the same Nvidia RTX A6000 GPU on the KITTI dataset, compared to the state-of-the-art one-frame method NeWCRFs \cite{yuan2022newcrfs}, the inference speed of our CTA-Depth, i.e., the number of images inferred per second (FPS), is greatly improved by 58.9$\%$, i.e., 5.53 (Ours) vs. 3.48 (NeWCRFs). This is because NeWCRFs use four swin-large transformers as multi-level encoders, while our method uses the lightweight ResNet18 backbone as the encoder to extract features. 
On the other hand, as shown in Table \ref{table_kitti_2}, although the FPS of CTA-Depth is slightly lower than that of the multi-frame method DRO \cite{gu2023dro} (5.53 vs. 6.25),  our performance significantly outperforms DRO and MaGNet \cite{bae2022multi}.

\subsection{Evaluation of Our Method}
% \subsection{Comparison with State-of-the-Arts}

\paragraph{Evaluation on KITTI.} 
We first compare our CTA-Depth against top performers of supervised monocular depth estimation on the KITTI dataset; see Tables \ref{table_kitti_1} \& \ref{table_kitti_2} for the results. For a fair comparison, all methods are evaluated given the same sequential images. %Tables~\ref{table_kitti_1} \& \ref{table_kitti_2} summarize the quantitative comparison results.
In Table \ref{table_kitti_1}, the seven widely-used evaluation metrics are calculated strictly following the work~\cite{gu2023dro} and the ground-truth median scaling is applied to obtain the final output.
% In contrast, the seven evaluation metrics in Table~\ref{table_kitti_2} are calculated according to AdaBins~\cite{bhat2021adabins} and the final output is calculated by taking the average prediction of the image and its mirror image. We also evaluate our method against others in terms of the reference speed (FPS) using the same Nvidia RTX A6000 GPU. Clearly, both CTA-Depth and CTA-Depth (L) achieve the state-of-the-art performance over all the evaluation metrics under two different evaluation strategies with significant improvements. 
In contrast, the seven evaluation metrics in Table~\ref{table_kitti_2} are calculated according to AdaBins~\cite{bhat2021adabins} and the final output is calculated by taking the average prediction of the image and its mirror image. 
We also evaluate our method against others in terms of the reference speed (FPS) using the same Nvidia RTX A6000 GPU. Clearly, CTA-Depth achieves state-of-the-art performance over all the evaluation metrics under two different evaluation strategies.

% Note that, "M$\rightarrow$O" means monocular multiple frame images are used in training while only one image is used for inference.Clearly, as shown in both two tables, our method (including two versions CTA-Depth and CTA-Depth (L)) achieves the highest values over all the evaluation metrics under two different calculation strategies with significant improvements (see the bottom-most row).
% ; see these regions marked by green boxes
We further show the qualitative comparisons in Fig.~\ref{fig005} by comparing our method (c) with the recent approach \cite{gu2023dro} (b).
As shown in the green boxes, our method yields finer depth estimation results for moving objects in dynamic scenes, small objects and object contours, such as the trams and traffic signs. In addition, as shown in the top row of Fig.~\ref{fig005}, our predicted depth map for the tram window is more consistent with the rest parts of the tram. 
\paragraph{Evaluation on Virtual KITTI 2.} We further verify our method on the virtual KITTI 2 dataset as shown in Table \ref{table_vkitti01}. 
We use a subset of the virtual KITTI 2, which contains 1,700 images for training and 193 images for testing. 
%The performance of our method compared with the most relevant method DRO \cite{gu2023dro} is shown in Table \ref{table_vkitti01}.
Notably, our CTA-Depth achieved significantly better results than the multi-frame baseline methods over all evaluation metrics.
% Notably, our CTA-Depth has achieved significantly better results than the multi-frame baseline methods over all evaluation metrics, even without the full pipeline CTA-Depth (L).
% \vspace{-2mm}
\paragraph{Evaluation on nuScenes.} To further demonstrate the competitiveness of our approach, we also conduct an evaluation on the nuScenes dataset. 
%Similarly, only the supervised learning scheme is used.
In this experiment, we manually split a subset consisting of 2,600 images for training and 170 images for testing. The result is shown in Table \ref{table_nus01}. Again, the results show that our proposed method outperforms the baselines with a significant margin in all evaluation metrics.
% \vspace{-2mm}
\begin{table}[t]
	\footnotesize
	\setlength{\tabcolsep}{3.5pt}
    % 	\vspace{-0.5em}
	% For LaTeX tables use
	\centering
	\begin{tabular}{lcccc}
		\toprule
		Method & Abs Rel $\downarrow$ & Sq Rel $\downarrow$ & RMSE $\downarrow$ & RMSE$_{log}$  $\downarrow$\\
		\midrule
		GLPDepth & 0.058 & 0.217 &  2.146 &  0.125 \\
		Adabins & 0.041 & 0.164 & 1.981 & 0.094 \\
		DRO & 0.040 & 0.153 &  1.903 &  0.092 \\ 
		\textbf{CTA-Depth}  & \textbf{0.035} & \textbf{0.129} &  \textbf{1.715} &  \textbf{0.085} \\
		\midrule
		% \textit{Improvement} & \textcolor{blue}{12.5$\%$} $\uparrow$ & \textcolor{blue}{15.7$\%$} $\uparrow$ &  \textcolor{blue}{9.9$\%$} $\uparrow$ &  \textcolor{blue}{7.6$\%$} $\uparrow$ \\
		\bottomrule
	\end{tabular}
	\caption{Quantitative results on a subset of the Virtual KITTI 2. }
 % Highlighting: \textbf{best}. , \textcolor{blue}{our improvement}
        \vspace{-3mm}
	\label{table_vkitti01}       % Give a unique label
\end{table}

\begin{table}[t]
	\footnotesize
	\setlength{\tabcolsep}{3.5pt}
    % 	\vspace{-0.5em}
	% For LaTeX tables use
	\centering
	\begin{tabular}{lcccc}
		\toprule
		Method             & Abs Rel $\downarrow$ & Sq Rel $\downarrow$ & RMSE $\downarrow$ & RMSE$_{log}$  $\downarrow$\\
		\midrule
		GLPDepth & 0.061 & 0.340 & 3.159 &  0.121 \\
		Adabins & 0.058 & 0.314 & 3.156 & 0.117\\	DRO &  0.057 & 0.303 &  3.004 &  0.114 \\
		\textbf{CTA-Depth}  &  \textbf{0.050} & \textbf{0.252} &  \textbf{2.786} &  \textbf{0.104} \\
		% \midrule
		% \textit{Improvement} & \textcolor{blue}{12.3$\%$} $\uparrow$ & \textcolor{blue}{16.8$\%$} $\uparrow$ &  \textcolor{blue}{7.3$\%$} $\uparrow$ &  \textcolor{blue}{8.8$\%$} $\uparrow$ \\
		\bottomrule
	\end{tabular}
	\caption{Quantitative results on a subset of the nuScenes dataset. }
	\vspace{-3mm}
	\label{table_nus01}       % Give a unique label
\end{table}

\subsection{Ablation Study}

\begin{table}[t]
	\footnotesize
	\setlength{\tabcolsep}{3.5pt}
	% For LaTeX tables use
	\centering
	\begin{tabular}{lccccc}
		\toprule
	Setting & Abs Rel $\downarrow$ & Sq Rel $\downarrow$ & RMSE $\downarrow$ & RMSE$_{log}$ $\downarrow$\\
% 	    \midrule
% 	w/o Cost & 0.065 & 0.324 & 3.270 & 0.112 & 0.940 & 0.988 \\
% 	Cost volume & 0.049 & 0.214 & 2.804 & 0.086 & 0.966 & 0.994 \\
% 	Baseline & 0.047 & 0.199 & 2.629 & 0.082 & 0.970 & 0.994 \\
% Baseline & 0.065 & 0.325 & 3.279 & 0.113 \\
%         + Refiner & 0.047 & 0.199 & 2.629 & 0.082 \\
%         + Depth-CTA & 0.044 & 0.178 & 2.463 & 0.077 \\
%         + Pose-CTA & 0.042 & 0.161 & 2.357 & 0.074 \\
% 	+ TEM & 0.040 & 0.157 & 2.288 & 0.071 \\
% 	+ M  & 0.040 & 0.155 & 2.276 & 0.071 \\
% 	+ M + C & 0.039 & 0.151 & 2.253 & 0.070 \\ 
% 	+ MAE  & 0.038 & 0.145 & 2.224 & 0.069 \\
	    \midrule
	Baseline & 0.060 & 0.275 & 3.132 & 0.104 \\
        + Multi-level & 0.059 & 0.267 & 3.105 & 0.101 \\
	+ Cross-scale & 0.057 & 0.248 & 2.951 & 0.096\\
	+ MAE  & 0.056 & 0.243 & 2.930 & 0.094 \\
        + Depth-CTA & 0.051 & 0.192 & 2.586 & 0.082 \\
        + Pose-CTA & 0.047 & 0.169 & 2.307 & 0.075 \\
	+ LGE & 0.045 & 0.156 & 2.275 & 0.073 \\
	
	% + M + E + R + P + S & 0.037 & 0.139 & 2.191 & 0.068 \\
 %        Baseline & 0.047 & 0.199 & 2.629 & 0.082 \\
	% CTA-Refiner & 0.040 & 0.157 & 2.288 & 0.071 \\
	% + M  & 0.040 & 0.155 & 2.276 & 0.071 \\
	% + M + C & 0.039 & 0.151 & 2.253 & 0.070 \\ 
	% + M + E + R & 0.039 & 0.148 & 2.245 & 0.070 \\
	% + M + E + R + P & 0.038 & 0.145 & 2.224 & 0.069 \\
	% + M + E + R + P + S & 0.037 & 0.139 & 2.191 & 0.068 \\
% 		\midrule\
% 	    16, 8, 4, 2 & 0.038 & 0.139 & 2.191 & 0.068 & 0.980 & 0.996 \\
% 		32, 16, 8, 4 & 0.037 & 0.139 & 2.191 & 0.068 & 0.980 & 0.996 \\
		\midrule
	$I_{t-1}$, $I_t$ & 0.050 & 0.221 & 2.813 & 0.089 \\
	$I_{t-2}$, $I_{t-1}$, $I_t$ & 0.048 & 0.215 & 2.807 & 0.086 \\
	$I_{t-1}$, $I_t$, $I_{t+1}$  & 0.045 & 0.156 & 2.275 & 0.073 \\
		\bottomrule
	\end{tabular}
	\caption{Ablation study on the KITTI dataset. "Multi-level" refers to multi-level feature extraction, "Cross-scale" refers to cross-scale attention layers, "CTA" refers to context-aware temporal attention. "$I_{t-i}$" is the input frame at time $t-i$.}
	\vspace*{-2mm}
	\label{tableAblationStudy}       % Give a unique label
\end{table}
%"Depth-CTA" and "Pose-CTA" refers to depth and pose context-aware temporal attention, respectively.Table \ref{tableAblationStudy} summarizes the ablation study results of each major module in our method on KITTI dataset.
To inspect the importance of each module in our method, we conduct an ablation study on the KITTI dataset and provide the results in Table \ref{tableAblationStudy}. From top to bottom, the proposed modules are added in turn until the full method is constructed.

\noindent\textbf{Baseline.} To verify the effectiveness of each component, we build a baseline model. 
This model has a similar network architecture as the full pipeline, which includes the encoder-decoder structure with a deep recurrent network. 
In other words, the proposed CTA module and LGM are removed.
The MAE module keeps only the single-level ResNet18 feature net for the depth and pose estimations.
% based on the baseline, we only replace its recurrent optimizer with our CTA-Refiner.and gain a significant performance improvement as shown in Table \ref{tableAblationStudy}. 
% Clearly, "Abs Rel" error is reduced from 0.047 to 0.040, demonstrating the effectiveness of the context-aware temporal attention based refinement network in depth estimation.

% we train our CTA-Depth model with the exactly same setting as the baseline DRO\cite{gu2023dro} model. In other words, 

% To verify the effectiveness of different components in multi-level attention enhanced predictor, we turn them on and off in turn as shown in Table \ref{tableAblationStudy}.

% \textbf{Multi-level feature extraction.} \lzz{Compared with baseline method, which only extract the feature maps with the 1/8 resolution of the input images, we extract four levels feature maps as shown in Fig.~\ref{fig:encoder}. The performance increment shows that multi-level features provide more valuable information.}

\noindent\textbf{Multi-level feature extraction.} We extract four levels of feature maps as shown in Fig.~\ref{fig:encoder}. The performance gain shows that it provides more valuable information for the model.

\noindent\textbf{Cross-scale attention.} Next, we use the same ResNet18 as the feature encoder and add cross-scale attention layers to decode the features at each level following Fig.~\ref{fig:encoder}.
%, which effectively maintains the global information interaction. 
% As shown in Table \ref{tableAblationStudy}, the cross-scale attention decoder leads to a performance increase.
%  Unlike DRO, the initial depth maps are obtained directly from the ResNet18 and a depth head, we use ResNet18 as the feature encoder and add cross-attention layers to decode the features at each level, which effectively maintains the global information interaction. 
% employ an encoder-decoder network to yield the initial predictions. the initial depth and pose predictions are obtained directly from the ResNet18 network and a depth head.In addition to the two operations above, the rearrange upscale is added to reduce the network complexity. Together with the PPM, they consequently enhance the estimation performance.

\noindent\textbf{MAE module.} In addition to the two operations above, the rearrange upscale is added to reduce the network complexity. Together with the PPM, they consequently enhance the estimation performance.
% \textbf{Swin-transformer encoder.} \lzz{In this step, we replace ResNet18 encoder with swin-transformer encoder and name this large network scheme as TA-Opt (L). For baseline and our TA-Opt, the main contribution is the optimization module. Since the optimization of depth and pose is a coarse-to-fine process, the initial predictions do not need to be very precise. Nevertheless, a stronger initial predictor still leads to an overall performance improvement of the framework.}

\noindent\textbf{Depth-CTA and Pose-CTA.} We add Depth-CTA and Pose-CTA after the cost map to obtain dynamic features by implicitly modeling the temporal relation and learn the geometric constraints through the temporal frames. In this way, the learned adaptive dynamic features are fed to the GRU optimizers and yield a noticeable performance gain. The "Abs Rel" error is reduced from 0.056 to 0.047. Furthermore, Fig.~\ref{fig:teaser} demonstrates the effectiveness of CTA modules in estimating accurate depths in dynamic environments.
%  Unlike DRO, the initial

\noindent\textbf{LGE.} The long-range geometry embedding module provides temporal priors for dynamic objects in several temporal neighboring frames and enhances the learning of the CTA module by large margins. 
% \noindent\textbf{Swin-transformer encoder.} In this step, we replace ResNet18 encoder with swin-transformer encoder and name this large network scheme as CTA-Depth (L), which also further improves the performance.

\noindent\textbf{Multi-frame input.} Here, we set up three experiments with different numbers of input frames. The results show that the optimal performance is achieved when the adopted temporal frames are $I_{t-1}$, $I_t$ and $I_{t+1}$, i.e., two image pairs: ($I_{t-1},I_{t}$) and ($I_{t},I_{t+1}$).
However, the network becomes over-complicated and time-consuming to train or infer when utilizing more than three sample frames for depth refinement.

\section{Conclusion}

In this work, we present a novel CTA-Depth for multi-frame monocular depth estimation. To resolve the ambiguity caused by challenging dynamic scenes with moving objects, we propose the CTA-Refiner by designing context-aware temporal attention to implicitly leverage temporal information and model image self-similarities. In addition, we develop a novel long-range geometry embedding module to efficiently inject our refiner with geometry reasoning among the long-range temporal frames.
Furthermore, we build a multi-level encoder-decoder network with the attention-enhanced predictor to obtain features with both global and local attentions. We achieve state-of-the-art performances on three challenging monocular depth estimation benchmarks.
In the future, we would like to employ our multi-frame association mechanism in relevant tasks such as 3D object detection.

\bibliographystyle{named}
\bibliography{ijcai23}

\end{document}